\documentclass[sigconf]{acmart}
\AtBeginDocument{%
  }

\copyrightyear{2026}
\acmYear{2026}
\setcopyright{cc}
\setcctype{by}
\acmConference[WWW '26] {Proceedings of the ACM Web Conference 2026}{April 13--17, 2026}{Dubai, United Arab Emirates.}
\acmBooktitle{Proceedings of the ACM Web Conference 2026 (WWW '26), April 13--17, 2026, Dubai, United Arab Emirates}
\acmISBN{979-8-4007-2307-0/2026/04}
\acmDOI{10.1145/3774904.3792144}
\usepackage{multirow}
\usepackage{algorithm}
\usepackage{algorithmic}
\usepackage{balance}

\begin{document}

\title[FedSRD: Communication-efficient Federated LLMs Finetuning]{FedSRD: Sparsify-Reconstruct-Decompose for Communication-Efficient Federated Large Language Models Fine-Tuning}

\author{Guochen Yan}
\affiliation{%
  \institution{Beijing Key Laboratory of Data Intelligence and Security, National Engineering Research Center for Software Engineering, School of Computer Science, Peking University}
  \city{Beijing}
  \country{China}
}
\email{2301111972@stu.pku.edu.cn}

\author{Luyuan Xie}
\affiliation{%
  \institution{Beijing Key Laboratory of Data Intelligence and Security, School of Software and Microelectronics, Peking University}
  \city{Beijing}
  \country{China}}
\email{2201110745@stu.pku.edu.cn}

\author{Qingni Shen}
\authornote{Corresponding author}
\affiliation{%
  \institution{Beijing Key Laboratory of Data Intelligence and Security, School of Software and Microelectronics, Peking University}
  \city{Beijing}
  \country{China}
}
\email{qingnishen@ss.pku.edu.cn}

\author{Yuejian Fang}
\affiliation{%
 \institution{Beijing Key Laboratory of Data Intelligence and Security, School of Software and Microelectronics, Peking University}
 \city{Beijing}
 \country{China}}
\email{fangyj@ss.pku.edu.cn}

\author{Zhonghai Wu}
\authornotemark[1]
\affiliation{%
  \institution{Beijing Key Laboratory of Data Intelligence and Security, National Engineering Research Center for Software Engineering, Peking University}
  \city{Beijing}
  \country{China}}
\email{wuzh@pku.edu.cn}

\renewcommand{\shortauthors}{Guochen Yan, Luyuan Xie, Qingni Shen, Yuejian Fang, and Zhonghai Wu}

\begin{abstract}
The current paradigm of training large language models (LLMs) on public available Web data is becoming unsustainable as high-quality data sources in specialized domains near exhaustion. Federated Learning (FL) emerges as a practical solution for the next generation of AI on a decentralized Web, enabling privacy-preserving collaborative fine-tuning on decentralized private data. While Low-Rank Adaptation (LoRA) is standard for efficient fine-tuning, its federated application faces a critical bottleneck: communication overhead under heterogeneous network conditions. Structural redundancy in LoRA parameters increases communication costs and causes aggregation conflicts. To address this, we propose FedSRD, a Sparsify-Reconstruct-Decompose framework for communication-efficient federated LLM fine-tuning. We introduce importance-aware sparsification to reduce the upload parameter count while preserving the structural integrity of LoRA updates. The server aggregates updates in full-rank space to mitigate conflicts, then decomposes the global update into a sparse low-rank format for broadcast, ensuring a symmetrically efficient cycle. We also propose an efficient variant, FedSRD-e, to reduce computational overhead. Experiments on 10 benchmarks show our framework significantly reduces communication costs by up to 90\% while improving performance on heterogeneous client data.
\end{abstract}




\maketitle

\section{Introduction}
Large Language Models (LLMs), pre-trained on vast web-scale corpora, have demonstrated remarkable generalist capabilities~\cite{ouyang2022training, kaplan2020scaling}. However, realizing their full potential for specialized downstream applications requires fine-tuning on diverse, high-quality, domain-specific data. Such data is often proprietary and distributed across numerous private sources, resulting in fragmented "data silos". As centralizing this sensitive data is frequently impractical due to privacy regulations and cost constraints, Federated Learning (FL) offers a compelling solution, enabling multiple clients to collaboratively fine-tune a global shared LLM without direct data exchange. This paradigm is central to building the next generation of intelligent, decentralized Web applications~\cite{sani2024future, wu2025survey}.

\begin{figure}[tbp]
    \centering
    \includegraphics[width=\linewidth]{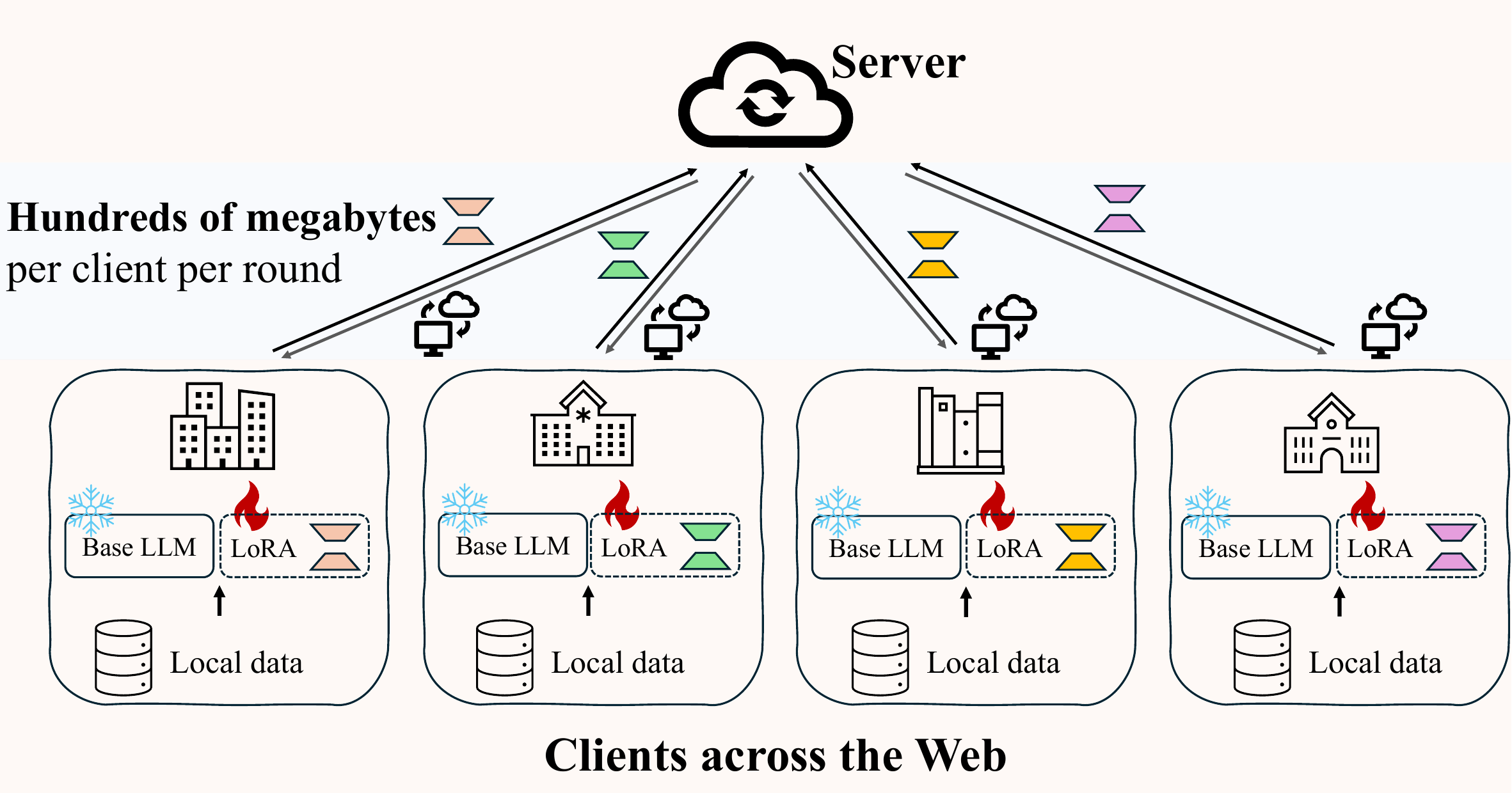}
    \caption{An illustration of Federated LoRA Fine-tuning. Clients train and communicate only the LoRA matrices. However, this payload can still amount to hundreds of megabytes per round, posing a significant bottleneck.}
    \label{fig:fl_demo}
\end{figure}

While FL provides a privacy-preserving solution, fine-tuning billion-parameter LLMs across distributed clients remains computationally and communicatively prohibitive. To overcome this, Parameter-Efficient Fine-Tuning (PEFT) methods, particularly Low-Rank Adaptation (LoRA)~\cite{hu2022lora}, have become the standard~\cite{ding2023parameter, kuang2024federatedscope, ye2024openfedllm}. As illustrated in Figure~\ref{fig:fl_demo}, LoRA significantly reduces the training burden by optimizing only a small set of trainable, low-rank matrices. In a federated setting, clients train and communicate only these LoRA parameters, achieving performance comparable to full-parameter fine-tuning at a fraction of the resource cost.

\begin{figure}[tbp]
    \centering
    \includegraphics[width=0.98\linewidth]{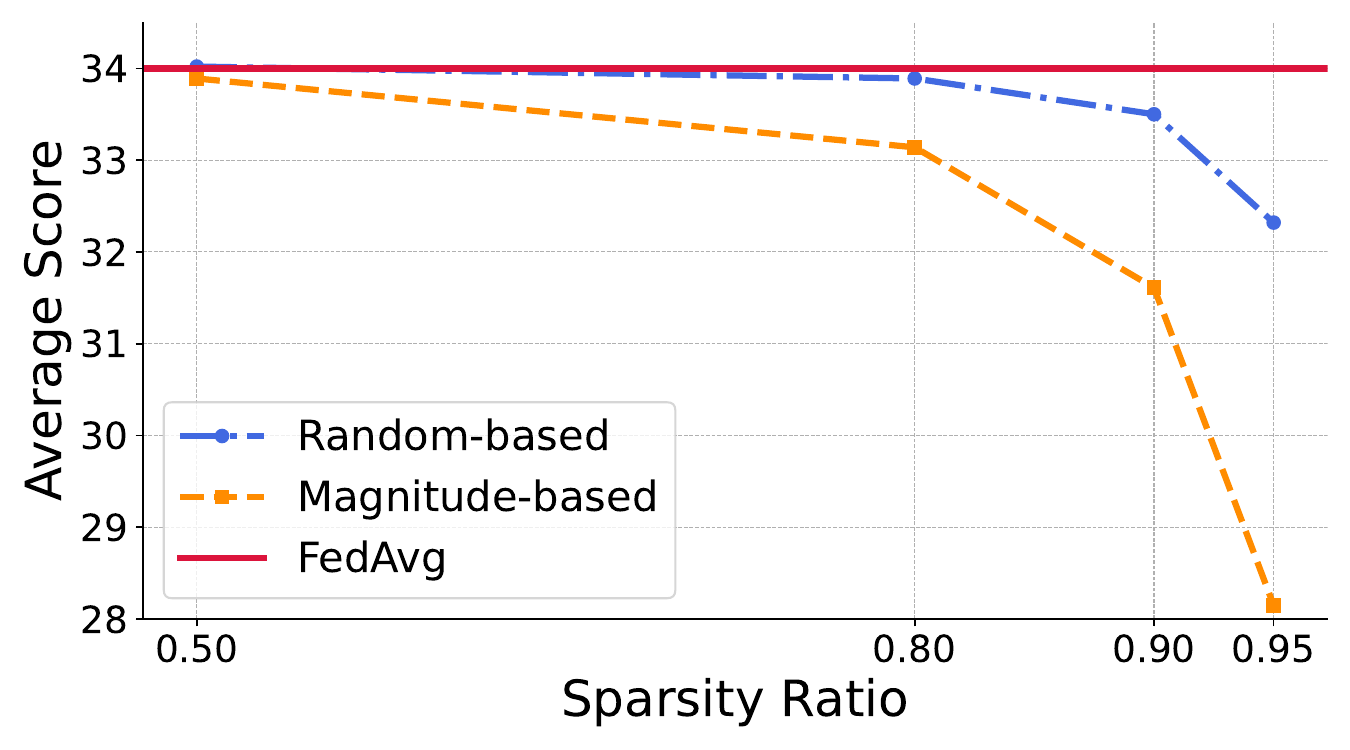}
    \caption{Performance comparison under different sparsity ratios. Both random and magnitude-based sparsification suffer performance decline at high sparsity ratios, motivating the need for a more advanced, structure-aware strategy.}
    \label{fig:sparse_method_comp}
\end{figure}

Despite the efficiency, the iterative communication of LoRA parameters remains a substantial overhead. This becomes a critical bottleneck when deployed across the heterogeneous and unpredictable network conditions characteristic of the global Web and mobile systems~\cite{dorfman2023docofl, mundt2004much}. High communication costs can exclude clients with limited bandwidth, such as those in rural areas or on mobile devices, undermining data diversity and posing fairness and inclusivity challenges for the global Web ecosystem. While sparsification is a common strategy for reducing communication, conventional methods are ill-adapted for federated LoRA fine-tuning on non-IID data. As shown in Figure~\ref{fig:sparse_method_comp}, naïve approaches like magnitude-based pruning~\cite{aji2017sparse} disrupt the structural relationship between LoRA matrices $A$ and $B$, while random sparsification~\cite{yu2024language} also degrades at high sparsity ratios. Crucially, both methods fail to preserve client-specific knowledge and exacerbate conflicts during server-side aggregation under data heterogeneity. This results in an undesirable trade-off: reduced communication at the expense of significant performance loss. This leads to our central research question:

\textbf{How can we significantly reduce communication costs in federated LLMs fine-tuning without sacrificing model performance under heterogeneous client data?}

To address this question, we introduce FedSRD, a novel Sparsify-Reconstruct-Decompose framework for communication-efficient federated LLMs fine-tuning. Our framework ensures high model performance through three sequential operations.

First, on the client side, we develop an importance-aware \textbf{Sparsification} strategy. Instead of relying on parameter magnitude, we calculate the importance scores that quantify the contribution of each element in update matrices ($\Delta B$ and $\Delta A$) to the overall update ($\Delta W$). This effectively preserves the critical structural interplay between LoRA matrices. Furthermore, recognizing that different layers serve distinct functions, we adaptively determine layer-wise sparsity ratios based on the kurtosis of its importance scores. Aggressive pruning is applied where high importance is concentrated. Second, the server executes the \textbf{Reconstruction} and \textbf{Decomposition}. Upon receiving sparse updates, the server first reconstructs each client's full-rank update ($\Delta W$) before aggregation. This enables aggregation to occur in the full-rank space, which is mathematically sound and effectively mitigates the update conflicts from non-IID data~\cite{wang2024flora, sun2024improving}. Finally, to create a symmetrically efficient communication cycle, the server performs a novel \textbf{Decomposition} of the aggregated global update back into a sparse, low-rank format for clients to download. Using a Taylor expansion-based approximation, we employ an alternating scheme that solves for and transmits only one sparse low-rank matrix per round. Furthermore, We propose FedSRD-e, a computational-efficient variant that bypasses intensive SVD operations for resource-constrained servers.

We demonstrate the effectiveness of FedSRD through extensive experiments that simulate practical non-IID scenarios by finetuning an LLM with clients from diverse specialist domains. Our evaluation on 8 in-domain and 2 out-of-domain benchmarks shows that FedSRD reduces both upload and download communication by 90\% while achieving even better performance. In summary, our contributions are:

\begin{itemize}
    \item We propose a novel, structure-aware sparsification method for federated LoRA fine-tuning based on importance scores. It effectively identifies and preserves the most salient parameters, overcoming the limitations of conventional sparsification methods.
    \item We propose a two-step server-side process: (1) reconstructing and aggregating client updates in a full-rank space to ensure mathematically sound aggregation and conflict mitigation, and (2) an alternating decomposition method to create sparse global updates for efficient download. We also introduce an efficient variant of this process to reduce server computation.
    \item We conduct extensive experiments demonstrating that our methods can save over 90\% of communication costs while achieving superior model performance, validating FedSRD as a practical and high-performing solution for federated LLMs fine-tuning on the Web.
\end{itemize}

\section{Related Works}

\subsection{Parameter-Efficient Fine-Tuning}
The large scale of modern LLMs makes the full-parameter fine-tuning computationally prohibitive in resource-constrained environments. To address this, Parameter-Efficient Fine-Tuning (PEFT) methods~\cite{ding2023parameter} freeze most of model parameters and only update a small part. Current approaches include adapter-based methods~\cite{he2021towards, edalati2025krona, houlsby2019parameter, hu2022lora} that insert small, trainable modules in models, and prompting methods~\cite{petrov2023prompting, lester2021power}, such as prefix-tuning~\cite{li2021prefix}, that add learnable tokens to the input sequence. Among these, Low-Rank Adaptation (LoRA)~\cite{hu2022lora} has emerged as a dominant and effective approach. The key insight of LoRA is that weight updates in fine-tuning possess a low intrinsic rank. LoRA represents each weight matrix update as the product of two smaller, trainable low-rank matrices, significantly reducing the number of trainable parameters and memory footprint while achieving comparable performance.

\subsection{Federated LLMs Fine-tuning}
Federated Learning (FL) enables the collaborative fine-tuning of LLMs on decentralized private data, addressing both data scarcity and privacy concerns. A common approach is to combine PEFT methods with standard aggregation algorithms~\cite{long2024dual, kuang2024federatedscope, zhang2024towards, nguyen2024flora, ye2024openfedllm}. Recent studies point out that this naive combination introduces aggregation inaccuracy, where separately averaging the decomposed LoRA matrices ($B$ and $A$) from clients is not mathematically equivalent to averaging the true underlying model updates ($BA$). Several advanced methods have been proposed to mitigate this issue. FFA-LoRA~\cite{sun2024improving} ensures a mathematically sound aggregation by freezing the LoRA $A$ matrices and only aggregating the $B$ matrices. FLoRA~\cite{wang2024flora} stacks clients' LoRA matrices to ensure accurate aggregation. FlexLoRA~\cite{bai2024federated} proposes to aggregate the true model updates ($BA$) on the server and then employs Singular Value Decomposition (SVD) to derive new global low-rank matrices for clients. FedICU~\cite{liaosplitting} disentangles client updates into consensus and divergence components, which are then aggregated separately.

\subsection{Communication-Efficient Federated LLMs Finetuning}
Communication overhead remains a primary bottleneck in FL systems~\cite{yan2025fedvck, yan2024towards, sattler2019robust}, a problem that is exacerbated when fine-tuning LLMs over the public internet. While PEFT methods provide a baseline level of efficiency, further optimizations are crucial. Structured methods impose constraints on the LoRA updates. For instance, FFA-LoRA~\cite{sun2024improving} halves the communication cost by transmitting only the $B$ matrices. Fed-SB~\cite{singhal2025fed} freezes the LoRA matrices entirely and instead inserts a small, trainable square matrix for aggregation. While these approaches reduce communication, they often do so at the cost of model performance by constraining the expressive capacity of LoRA. Other lines of work focus on generic compression. Quantization methods~\cite{aji2017sparse, konevcny2016federated} represent parameters with fewer bits, but they often fail to achieve high compression ratios and can lead to significant accuracy degradation, particularly in non-IID settings. Sparsification methods, which transmit only a subset of non-zero parameters, can achieve much higher compression ratios~\cite{aji2017sparse, sahu2021rethinking}. Yet, traditional approaches are not directly applicable to LoRA. Magnitude-based pruning~\cite{aji2017sparse} is suboptimal because it ignores the coupled relationship between the two LoRA matrices. Similarly, random sparsification~\cite{yang2025impart, yu2024language} fails to preserve critical parameters, resulting in inferior performance at high sparsity ratios.

\section{Preliminaries}

\subsection{LoRA in LLMs Fine-tuning}
LoRA has been proven to exhibit better performance than other PEFT methods in the federated learning system~\cite{kuang2024federatedscope}. LoRA constrains the weight update in low-rank space:
\begin{equation}
    W_0 + \Delta W = W_0 + BA, 
\end{equation}
where $W_0$ is the frozen model weights, $B \in \mathbb R^{d_{out} \times r}$ and $A\in \mathbb R^{r \times d_{in}}$ are paired low-rank trainable matrices inserted in the linear weights to approximate the update $\Delta W \in \mathbb R^{d_{out} \times d_{in}}$, where $r \ll \min(d_{out}, d_{in})$. $B$ is initialized to zero and $A$ is initialized with random Gaussian noises~\cite{hu2022lora}.

\subsection{Problem Formulation}
We consider an FL system with one server and $m$ clients. Each client has its local private dataset $\mathcal D^i = \{x_j^i, y_j^i\}_{j=1}^{|\mathcal D^i|}$. These datasets may originate from different domains or tasks, creating a heterogeneous (non-IID) scenario. All clients collaboratively fine-tune a shared LLM $\mathcal{M}$ to perform well on their respective data distributions. This can be formulated as the following optimization problem:
\begin{equation}
    \min_{\mathcal M} \mathcal{L}(\mathcal M) = \sum_i^m \mathcal{L}_i(\mathcal M, \mathcal D^i),
\end{equation}
where $\mathcal{L}_i(\mathcal M, \mathcal D^i)$ represents the loss of the LLM $\mathcal{M}$ on local dataset $\mathcal{D}^i$. Following the recent studies, we adopt LoRA to fine-tune the LLM $\mathcal{M}$ in local fine-tuning on each client. The local optimization of client $i$ could be formulated as:
\begin{equation}
    \min \mathcal{L}_i = \min_{\mathcal P(\mathcal M)} \mathbb E_{(x_j^i, y_j^i) \sim \mathcal D^i} L(x_j^i, y_j^i, \mathcal P(\mathcal M)),
\end{equation}
where $\mathcal{P}(\mathcal M)$ denotes the trainable parameters set of LoRA. $L$ is the cross-entroy loss on completion $y_j^i$ given $x_j^i$.

\section{Proposed Method}
\begin{figure*}
    \centering
    \includegraphics[width=\textwidth]{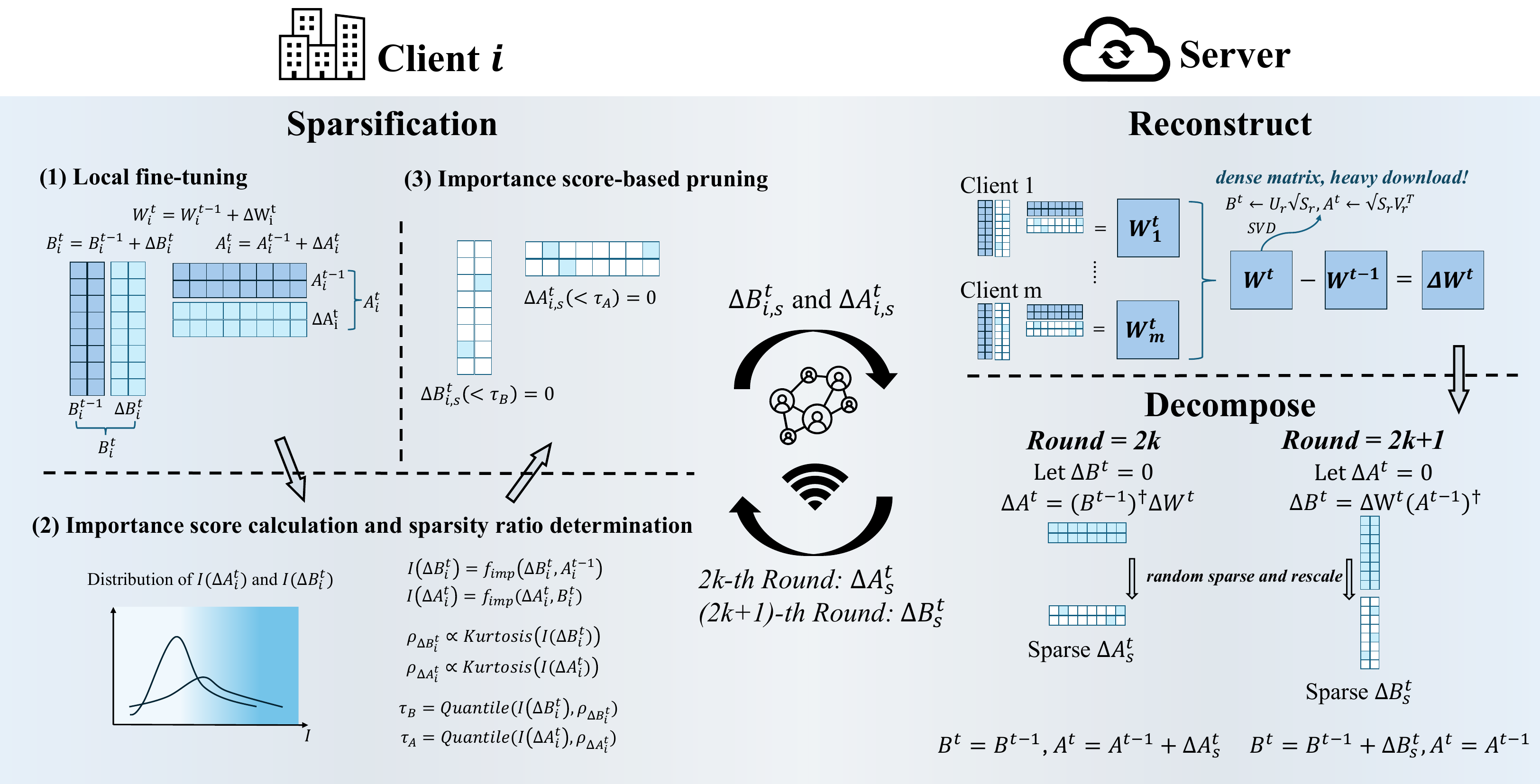}
    \caption{The FedSRD framework. (1) Client-Side sparsification: Each client computes its LoRA updates $\Delta B_i^t$ and $\Delta A_i^t$ and uses our importance-aware sparsification to generate sparse updates $\Delta B_{i, s}^t$ and $\Delta A_{i, s}^t$ for upload. (2) Server-Side reconstruction to aggregation: The server reconstructs each client's full weight matrix $W_i^t$ and aggregates them in the full-rank space. (3) Server-Side decomposition: The server computes the global update $\Delta W^t$, which is then decomposed into a single sparse matrix for efficient download.}
    \label{fig:framework}
\end{figure*}

In this section, we introduce FedSRD, a noval Sparsify-Reconstruct-Decompose framework for federated LoRA fine-tuning, illustrated in Figure~\ref{fig:framework}. At each communication round $t$, a client $i$ first computes its local LoRA updates $\Delta B_i^t$ and $\Delta A_i^t$, then uses our proposed importance-aware strategy to sparsify them for upload (Section~\ref{sec:importance_aware_sparse}). The server then reconstructs each client's full weight matrix and aggregates them in the full-rank space to mitigate non-IID conflicts (Section~\ref{sec:reconstruct}). Finally, to ensure an efficient download, the server decomposes the resulting global update into a single sparse low-rank matrix for broadcast using an alternating, Taylor approximation-based method (Section~\ref{sec:alternative_decompose}).

\subsection{Importance-aware Sparsification}
\label{sec:importance_aware_sparse}

\paragraph{Motivation} To reduce upload communication costs, sparsification is a natural approach. However, mainstream methods are ill-suited for federated LoRA fine-tuning. Magnitude-based sparsification is ineffective as it prunes each LoRA matrix independently, ignoring the crucial structural interplay between $B$ and $A$. On non-IID data, random-based sparsification fails to preserve performance at high sparsity ratios because it cannot differentiate between critical and unimportant parameters. This motivates a sparsification method that can (1) account for the structural relationship between $B$ and $A$ and (2) robustly preserve the most salient parameters.

To address these challenges, we propose importance-aware sparsification. Instead of transmitting the full LoRA matrices, we transmit only their updates at each round. Specifically, on client $i$ at round $t$, we first conduct LoRA fine-tuning on its local dataset to obtain the new LoRA matrices $B_i^t$ and $A_i^t$, where
\begin{equation}
    B_i^t = B_i^{t-1} + \Delta B_i^{t}, \quad A_i^t = A_i^{t-1} + \Delta A_i^{t},
\label{eq:lora_recover}
\end{equation}
where $B_i^{t-1}$ and $A_i^{t-1}$ are the matrices from the previous round~\footnote{Since our operations are layer independent, we omit layer-specific notations for clarity.}. The full-rank weight update $\Delta W_i^t$ can be expressed as:
\begin{equation}
\begin{aligned}
    \Delta W_i^t &= B_i^t A_i^t - B_i^{t-1}A_i^{t-1} \\
    &=B_i^t(A_i^{t-1} + \Delta A_i^t) - B_i^{t-1}A_i^{t-1} \\
    &= (B_i^{t} - B_i^{t-1})A_i^{t-1} + B_i^t\Delta A_i^{t-1} \\
    &= \Delta B_i^t A_i^{t-1} + B_i^t \Delta A_i^t.
\end{aligned}
\label{eq:local_update}
\end{equation}
This formulation reveals that the true contribution of an element in $\Delta B_i^t$ depends on the corresponding row in $A_i^{t-1}$, and likewise for $\Delta A_i^t$ and $B_i^t$. To formalize this, we therefore define importance score $I(\cdot)$ to measure an element's contribution to the final weight update by scaling its magnitude with the norm of the corresponding vector it interacts with in the other matrix:
\begin{equation}
\begin{aligned}
    I(\Delta B_i^t[u, v]) &= |\Delta B_i^t[u, v]| \cdot \|A_i^{t-1}[v, :]^T\|_2, \\
    I(\Delta A_i^t[u, v]) &= |\Delta A_i^t[u, v]| \cdot \|B_i^{t}[:, u]||_2.
\end{aligned}
\label{eq:importance_score}
\end{equation}
A higher importance score indicates a greater contribution to the final weight update. We then prune elements with the lowest scores. We explain the rationale behind and prove that pruning based on this score minimizes the upper bound of the weight update approximation error $\|\Delta W_i^t\|_F$ in Appendix~\ref{sec:rationale_imp}. 

Furthermore, recognizing that LoRA matrices in different layers play distinct roles~\cite{luo2025one}, we adaptively assign a sparsity ratio $\rho$ to each matrix. The core intuition is that the kurtosis of the importance score reveals a layer's functional "prunability". A high kurtosis (a leptokurtic, heavy-tailed distribution) signifies that importance is concentrated in a few parameters, justifying an aggressive sparsity ratio. Conversely, a low kurtosis (a platykurtic, flat distribution) implies that importance is diffused, necessitating a more conservative ratio. The layer-specific sparsity ratio is thus calculated as:
\begin{equation}
\begin{aligned}
    \rho_{\Delta B_i^t} &= \alpha + 0.1 \cdot \log(\text{kurtosis}(I(\Delta B_i^t))), \\
    \rho_{\Delta A_i^t} &= \alpha + 0.1 \cdot \log(\text{kurtosis}(I(\Delta A_i^t))), \\
\end{aligned}
\label{eq:kurtosis}
\end{equation}
where $\alpha$ is a hyperparameter. The sparsity ratios are truncated if they exceed a pre-set upper bound to ensure validity. We provide a detailed explanation and justification of this design in Appendix~\ref{sec:justification_eq7}. We use the adaptive sparsity ratios to prune $\Delta B_i^t$ and $\Delta A_i^t$:
\begin{equation}
\begin{aligned}
(\Delta B_{i, s}^t)_{uv} &=
\begin{cases}
    (\Delta B_i^t)_{uv} & \text{if } I(\Delta B_i^t)_{uv} > \text{quantile}(I(\Delta B_i^t), \rho_{\Delta B_i^t}) \\
    0 & \text{otherwise}
\end{cases} \\
(\Delta A_{i, s}^t)_{uv} &=
\begin{cases}
    (\Delta A_i^t)_{uv} & \text{if } I(\Delta A_i^t)_{uv} > \text{quantile}(I(\Delta A_i^t), \rho_{\Delta A_i^t}) \\
    0 & \text{otherwise}
\end{cases},
\end{aligned}
\end{equation}
where the $\Delta B_{i, s}^t$ and $\Delta A_{i, s}^t$ are sparsified matrices for uploading. The client can upload the sparse matrices in a compressed format.

\subsection{Reconstruct to Aggregation}
\label{sec:reconstruct}

\paragraph{Motivation} Most existing federated learning methods like FedAvg~\cite{mcmahan2017communication}, when applied to LoRA, average the clients' LoRA matrices independently. This aggregation is not mathematically sound and introduces unexpected noises because its multiplicative relationship between the matrices and fails to mitigate update conflicts arising from non-IID client data~\cite{sun2024improving, bai2024federated, wang2024flora}. This motivates us to develop an advanced aggregation method.

To enable a mathematically sound aggregation that preserves client knowledge, we propose to first reconstruct each client's update in the full-rank space. At round $t$, upon receiving the sparse updates $\Delta B_{i, s}^t$ and $\Delta A_{i, s}^t$ from $m$ participating clients, the server first computes each client's new LoRA matrices, $B_i^t$ and $A_i^t$, using Equation~\ref{eq:lora_recover}. It then reconstructs each client's full-rank weight matrix and aggregates them:
\begin{equation}
    W^t = \frac{1}{m}\sum_i^m B_i^t A_i^t,
\end{equation}
where $W^t$ is the aggregated LoRA weights in the full-rank parameter space. By performing aggregation in this higher-dimensional space, the client-specific knowledge encoded in $B_i^t$ and $A_i^t$ is properly combined, reducing the likelihood of destructive interference.

\subsection{Alternative Decomposition}
\label{sec:alternative_decompose}
\paragraph{Motivation} Previous works~\cite{bai2024federated, lee2025fedsvd} often use Singular Value Decomposition (SVD) to decompose aggregated weights into new dense matrices $B^t$ and $A^t$. But transmitting them creates a download bottleneck. Naively sparsifying the full state matrices $B^t$ and $A^t$ would damage the learned model. Computing the update as a direct difference, $B^t - B^{t-1}$, is also mathematically flawed. This is because SVD is rotation-invariant and decomposition is not unique. Even a small global update can induce an arbitrary rotation of the low-rank basis between rounds, creating large, dense updates that do not reflect the true minimal updates. Thus, they are highly unsuitable for sparsification. These motivate the need for a novel decomposition capable of producing a stable update for sparse downloads.

Our goal is to find sparse updates $\Delta B^t$ and $\Delta A^t$ that approximate the change from the previous global state $W^{t-1}$ to the new one $W^t$. First, we project $W^t$ to its $r$-rank approximation via SVD:
\begin{equation}
\begin{aligned}
    U, S, V_h &= SVD(W^t), \\
    W^t_{low\_rank} &= U[:, :r] S[:r, :r] V_h[:r, :],
\end{aligned}
\end{equation}
where $W^t_{low\_rank}$ is the $r$-rank approximation of $W^t$, $r$ is the LoRA rank. This process inherently denoises the aggregated result by discarding less important components and retaining the consensus. Assuming the global states are low-rank, we can state that $\exists B^{t-1}, A^{t-1}$ such that $W^{t-1}_{\text{low\_rank}} = B^{t-1}A^{t-1}$, and we seek $\Delta B^t, \Delta A^t$ such that $W^t_{\text{low\_rank}} = (B^{t-1} + \Delta B^t)(A^{t-1} + \Delta A^t)$. We can approximate the update using a first-order Taylor expansion:
\begin{equation}
\begin{aligned}
    W&^t_{low\_rank} - W^{t-1}_{low\_rank} \\
    &= (B^{t-1} + \Delta B^t)(A^{t-1} + \Delta A^t) - B^{t-1}A^{t-1} \\
    &= B^{t-1}\Delta A^{t} + \Delta B^t A^{t-1} + \Delta B^t \Delta A^t \\
    &\approx B^{t-1}\Delta A^{t} + \Delta B^t A^{t-1}, 
\end{aligned}
\label{eq:taylor_expension}
\end{equation}
where the higher-order term $\Delta B^t \Delta A^t$ is omitted (discussed in Appendix~\ref{sec:app_magnitude_ratio}). To solve this underdetermined system, we alternate between setting one of the deltas to zero. In even-numbered rounds, we set $\Delta B^t = \mathbf{0}$ and solve for $\Delta A^t$:
\begin{equation}
\begin{aligned}
    &W^t_{low\_rank} - W^{t-1}_{low\_rank} = B^{t-1}\Delta A^{t}, \\
    &\Delta A^t = (B^{t-1})^{\dagger}(W^t_{low\_rank} - W^{t-1}_{low\_rank}),
\end{aligned}
\end{equation}
where $(B^{t-1})^{\dagger}$ is the Moore-Penrose pseudoinverse. In odd-numbered rounds, we set $\Delta A^t = \mathbf{0}$ and solve for $\Delta B^t$:
\begin{equation}
\begin{aligned}
    &W^t_{low\_rank} - W^{t-1}_{low\_rank} = \Delta B^t A^{t-1}, \\
    &\Delta B^t = (W^t_{low\_rank} - W^{t-1}_{low\_rank})(A^{t-1})^{\dagger},
\end{aligned}
\end{equation}
The resulting single update matrix ($\Delta A^t$ or $\Delta B^t$) can be simply sparsified using random pruning and rescaled~\cite{yu2024language} to produce a sparse update ($\Delta A^t_s$ or $\Delta B_s^t$) for download. Clients receive this single sparse matrix and update their local model accordingly, completing the communication-efficient cycle.

\paragraph{\textbf{FedSRD-e: a Computationally Efficient Variant}}
\label{sec:FedSRD-e}
For scenarios where server-side computation is the primary bottleneck, such as in edge devices, we propose a computationally efficient variant. This variant, FedSRD-e, bypasses the SVD projection and instead directly approximates the full-rank global update, $\Delta W^t = W^t - W^{t-1}$. This matrix $\Delta W^t$ replaces $W^t_{\text{low\_rank}} - W^{t-1}_{\text{low\_rank}}$ on the left-hand side of Equation~\ref{eq:taylor_expension}. The subsequent steps of solving for a single update matrix remain the same. This variant significantly reduces computation by avoiding the costly SVD operation. It only consumes slightly more time than the simple FedAvg method, serving as a powerful, high-speed alternative.

\paragraph{Remark} Note that both the reconstruction and decomposition operations are layer-independent, meaning they can be performed layer by layer. Thus, the additional memory resource demand on the server is negligible.

\begin{table}[tbp]
    \centering
    \caption{Details of the local datasets used by each client. All datasets are from the training split.}
    \label{tab:datasets_task}
    \resizebox{0.9\linewidth}{!}{
        \begin{tabular}{lll}
            \toprule
            Domain & Dataset & Task Type \\
            \midrule
            Code & Code Alpaca & Code Generation \\ \cmidrule(r){1-3}
            Medical & MedQA & Medical Question Answering \\ \cmidrule(r){1-3}
            Finance & Finance Alpaca & Financial Question Answering \\ \cmidrule(r){1-3}
            \multirow{2}{*}{Math} & GSM8K (train) & Mathematical Reasoning \& \\ 
             & Math (train) & Math Problems Solving \\
            \bottomrule
        \end{tabular}
    }
\end{table}

\begin{table}[tbp]
    \centering
    \caption{Details of the benchmarks used in our evaluation.}
    \label{tab:domain_benchmark}
    \resizebox{\linewidth}{!}{
        \begin{tabular}{llccc}
            \toprule
            Category & Domain & Benchmark & \# Instance & $k$-Shot \\ 
            \midrule
            \multirow{8}{*}{In-domain} & \multirow{2}{*}{Code} & HumanEval~\cite{chen2021evaluating} & 164 & 0 \\ 
             &  & Sanitized MBPP~\cite{austin2021program} & 257 & 3 \\ 
            \cmidrule(l){2-5}
             & \multirow{2}{*}{Medical} & MedQA~\cite{jin2021disease} & 1,273 & 1 \\
             &  & MedMCQA~\cite{pal2022medmcqa} & 4,183 & 1 \\ 
            \cmidrule(l){2-5}
             & \multirow{2}{*}{Finance} & FinEval~\cite{zhang2023fineval} & 4,661 (34 sub-tasks) & 0 \\ 
             &  & FinanceIQ~\cite{zhang2023xuanyuan} & 7,173 (10 sub-tasks) & 0 \\ 
            \cmidrule(l){2-5}
             & \multirow{2}{*}{Math} & GSM8K~\cite{cobbe2021training} & 1319 & 0 \\ 
             &  & MATH~\cite{hendrycks2021measuring} & 5000 & 0 \\ 
            \midrule
            \multirow{2}{*}{Out-of-domain} & General & AGIEval~\cite{zhong2023agieval} & 8,062 (20 sub-tasks) & 0 \\ 
             & Law & LawBench~\cite{fei2023lawbench} & 10,000 (20 sub-tasks) & 1 \\
            \bottomrule
        \end{tabular}
    }
\end{table}

\section{Experiments}

\subsection{Experimental Setup}

\subsubsection{Datasets}
We simulate an FL system with 4 clients from distinct specialist domains: Code, Medical, Finance, and Math. We use CodeAlpaca~\cite{codealpaca}, the training set of MedQA~\cite{jin2021disease}, Finance Alpaca\footnote{https://huggingface.co/datasets/gbharti/finance-alpaca}, and a combination of the training sets from GMS8K~\cite{cobbe2021training} and MATH~\cite{hendrycks2021measuring} as the local datasets for these clients, respectively.

\subsubsection{Baselines}
We compare our framework FedSRD and FedSRD-e with 6 state-of-the-art federated learning methods: (1) FedAvg~\cite{mcmahan2017communication}, (2) FedProx~\cite{li2020federated}, (3) FFA-LoRA~\cite{sun2024improving}, which freezes LoRA $A$ matrcies to ensure efficient communication and effective aggregation, (4) FlexLoRA~\cite{bai2024federated}, which aggregates in full-rank space and uses SVD for download, (5) FedICU~\cite{liaosplitting}, which selectively updates on global model and perform consensus-divergence splitting in the server-side aggregatio, (6) Fed-DARE, which applies random sparsification with rescaling (DARE~\cite{yu2024language}) to LoRA updates for upload.

\subsubsection{Benchmarks}
We evaluate baselines and our methods on 10 benchmarks, including 8 in-domain benchmarks from Code, Medical, Finance, and Math domains, and 2 out-of-domain benchmarks to evaluate the knowledge retention of other domains during fine-tuning. We provide details of these benchmarks in Table~\ref{tab:domain_benchmark}.

\subsubsection{Implementation}
In our experiments, we adopt Llama3.2-3B~\footnote{https://huggingface.co/meta-llama/Llama-3.2-3B} and Qwen2-7B~\cite{team2024qwen2} as base models for federated fine-tuning over 50 rounds. We apply LoRA to all linear layers with rank $r=64$ for Llama3.2-3B and $r=32$ for Qwen2-7B. The learning rate is tuned from 1e-5 to 2e-4. For Fed-DARE, the sparsity ratio is 0.9 as suggested~\cite{yu2024language}. For our method, we tune $\alpha$ in Equation~\ref{eq:kurtosis} within [0.85, 0.9] and set the download sparsity ratio to 0.8. For other hyperparameters, we follow the suggested values from the original papers. We report average performance on in-domain and out-of-domain benchmarks, as well as the per-round communication cost per client. We conduct training on NVIDIA A800 GPUs and inference on NVIDIA GeForce RTX 3090 GPUs with bfloat16 precision.

\begin{table*}[tbp]
\centering
\caption{Comparison of model performance and per-round communication cost. The top section uses Llama3.2-3B and the bottom section uses Qwen2-7B. OOD represents Out-of-Domain. Best and second-best results are bolded and underlined.}
\label{tab:main_results}
\resizebox{\textwidth}{!}{%
\begin{tabular}{@{}l c c c | c c c c c c c c | c c @{}}
\toprule[1pt]
\multirow{3}{*}{Method} & \multirow{3}{*}{\begin{tabular}{@{}c@{}}\textbf{Comm.}\\\textbf{Cost}\\\textbf{(MB)}\end{tabular}} & \multicolumn{2}{c|}{\textbf{Average}} & \multicolumn{8}{c|}{In-Domain} & \multicolumn{2}{c}{Out-of-Domain} \\
\cmidrule(lr){3-4} \cmidrule(lr){5-12} \cmidrule(lr){13-14}
& & \multirow{2}{*}{\textbf{In-Domain}} & \multirow{2}{*}{\textbf{OOD}} & \multicolumn{2}{c}{Code} & \multicolumn{2}{c}{Medical} & \multicolumn{2}{c}{Finance} & \multicolumn{2}{c|}{Math} & \multirow{2}{*}{AGIEval} & \multirow{2}{*}{Law} \\
\cmidrule(lr){5-6} \cmidrule(lr){7-8} \cmidrule(lr){9-10} \cmidrule(lr){11-12}
& & & & HumanEval & MBPP & MedQA & MedMCQA & FinEval & FinanceIQ & GSM8K & MATH & & \\

\midrule
\multicolumn{14}{c}{\textbf{Base Model: Llama3.2-3B}} \\
\midrule
FedAvg & 742 & 33.90 & 24.36 & 26.85 & 47.86 & 40.45 & \underline{41.53} & 41.96 & 38.08 & \underline{28.96} & 5.56 & 22.47 & 26.24 \\
FedProx & 742 & 33.43 & 23.82 & 28.05 & \underline{48.25} & 35.98 & 40.06 & \underline{42.75} & 37.82 & 28.05 & \underline{5.70} & 21.58 & 26.06 \\
FFA-LoRA & 371 & 28.39 & 26.12 & 3.05 & 46.30 & 39.95 & 40.11 & 40.23 & 36.02 & 24.79 & 5.10 & 24.63 & \textbf{27.60} \\
FlexLoRA & 742 & 32.36 & 26.08 & 28.75 & 47.47 & 36.29 & 41.24 & 42.22 & 37.09 & 27.37 & 5.68 & 24.92 & \underline{27.24} \\
FedICU & 568 & 33.91 & 25.76 & \textbf{31.10} & \textbf{49.42} & 40.15 & 41.07 & 40.83 & 37.56 & 25.09 & 5.26 & 24.77 & 26.75 \\
Fed-DARE & 190 & 33.57 & 24.25 & 26.60 & \underline{48.25} & 39.28 & 40.04 & 40.40 & 38.04 & 26.05 & 5.26 & 21.94 & 26.56 \\ \midrule
\textbf{FedSRD-e} & \textbf{74} & \underline{34.26} & \textbf{27.95} & \underline{29.27} & 45.87 & \underline{41.63} & 41.17 & 42.22 & \textbf{39.47} & \textbf{30.48} & \textbf{5.88} & \textbf{28.81} & 27.08 \\
\textbf{FedSRD} & \textbf{74} & \textbf{35.12} & \underline{26.17} & \textbf{31.10} & \textbf{49.42} & \textbf{41.95} & \textbf{41.62} & \textbf{43.44} & \underline{39.13} & \underline{28.96} & 5.36 & \underline{25.54} & 26.79 \\

\midrule
\multicolumn{14}{c}{\textbf{Base Model: Qwen2-7B}} \\
\midrule
FedAvg & 616 & 57.04 & 41.02 & 59.15 & \underline{37.35} & 61.67 & 75.23 & 65.16 & 49.04 & 64.29 & 44.46 & 48.25 & 33.79 \\
FedProx & 616 & 58.27 & 40.72 & 63.42 & 36.58 & 62.13 & 77.34 & 65.77 & 49.96 & 66.26 & 44.72 & \textbf{48.62} & 32.82 \\
FFA-LoRA & 308 & 53.36 & \textbf{43.41} & \textbf{68.29} & 17.90 & 51.55 & 57.02 & \textbf{67.76} & 48.92 & 69.37 & \textbf{46.08} & 46.92 & \textbf{39.89} \\
FlexLoRA & 616 & 55.89 & 40.49 & 63.41 & 35.02 & 58.83 & 68.30 & \underline{67.68} & \underline{50.53} & 58.61 & 44.70 & 47.81 & 33.17 \\
FedICU & 472 & 57.98 & 42.56 & 60.37 & \textbf{38.91} & 60.33 & 78.72 & 65.86 & 48.83 & 66.19 & 44.62 & \underline{48.44} & 36.28 \\
Fed-DARE & 158 & 57.01 & 41.72 & 60.37 & 31.91 & 61.12 & 75.30 & 66.03 & \textbf{50.84} & 65.96 & 44.58 & 48.37 & 35.07 \\ \midrule
\textbf{FedSRD-e} & \textbf{61} & \underline{61.13} & \underline{42.59} & \underline{68.09} & 36.56 & \textbf{77.80} & \underline{78.10} & 64.21 & 48.90 & \textbf{70.20} & 45.20 & 45.82 & \underline{39.35} \\
\textbf{FedSRD} & \textbf{61} & \textbf{61.19} & 42.58 & \textbf{68.29} & 36.19 & \underline{77.77} & \textbf{78.94} & 64.55 & 48.48 & \underline{69.93} & \underline{45.36} & 45.93 & 39.22 \\
\bottomrule[1pt]
\end{tabular}
}
\end{table*}

\begin{figure}[tbp]
    \centering
    \includegraphics[width=0.98\linewidth]{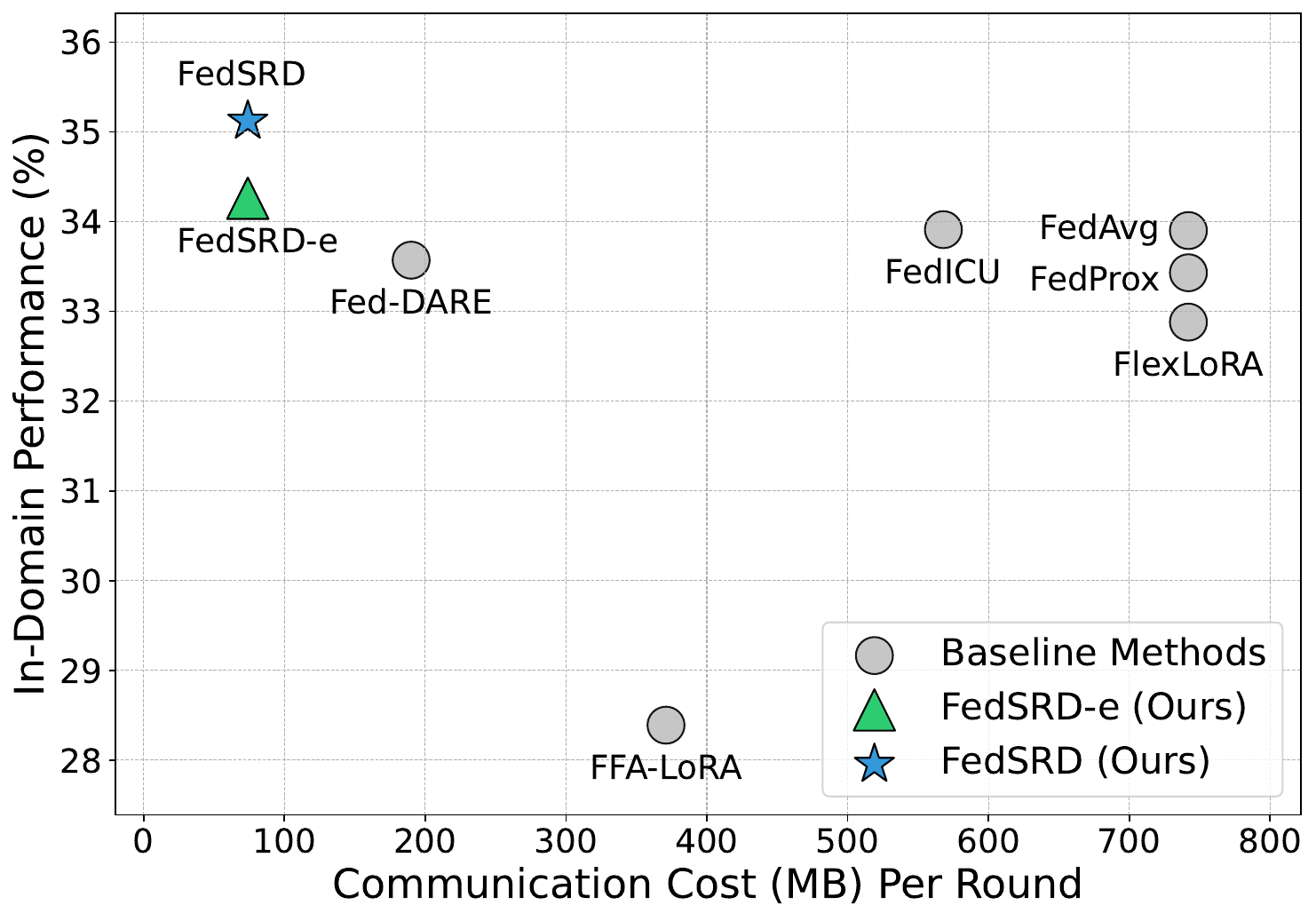}
    \caption{Average in-domain performance vs. per-round communication cost (Llama3.2-3B). Our methods are Pareto optimal, achieving the highest performance with the lowest communication cost.}
    \label{fig:perf_cost_time_llama3.2_3b}
\end{figure}

\begin{figure}[tbp]
    \centering
    \includegraphics[width=0.98\linewidth]{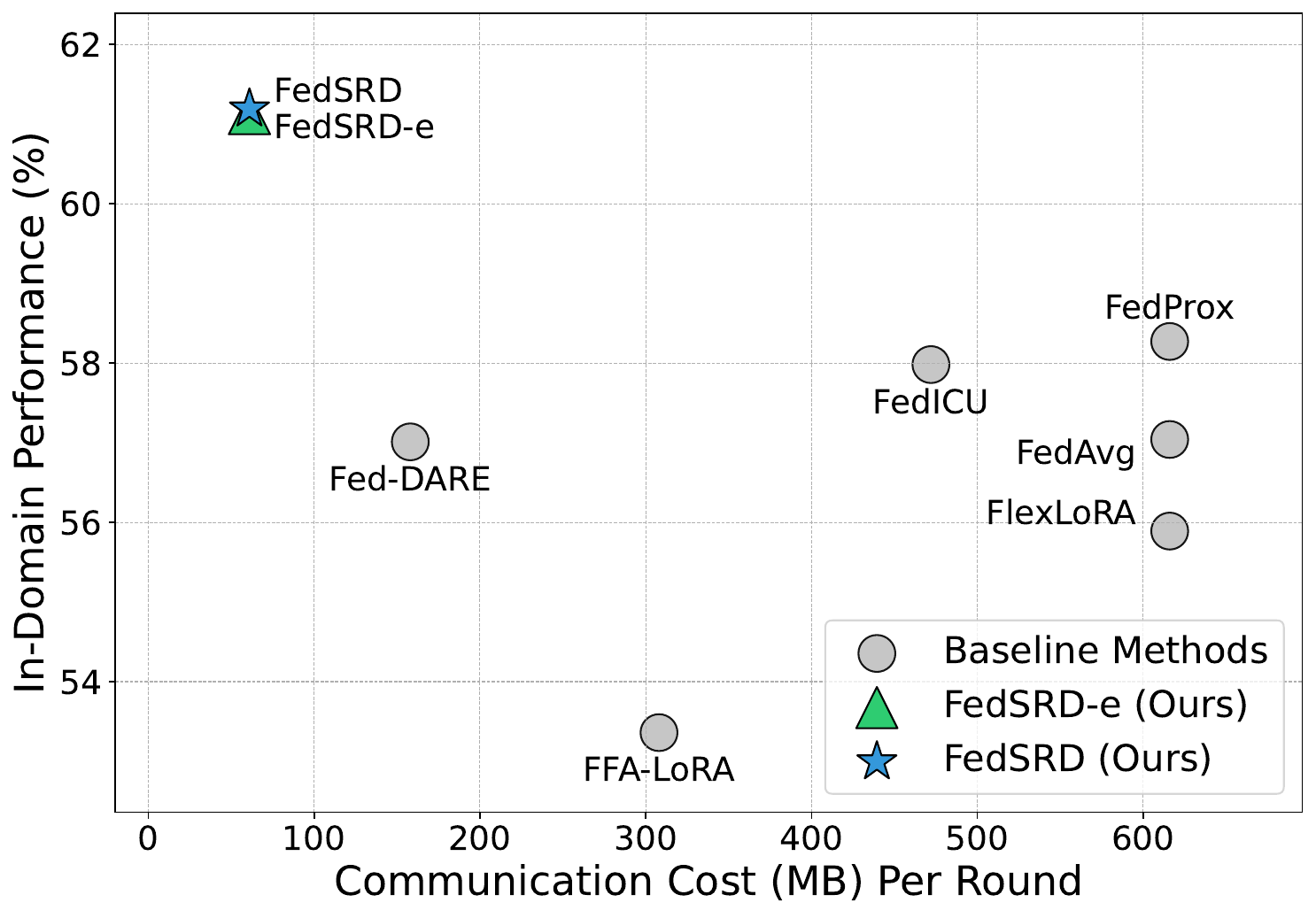}
    \caption{Average in-domain performance vs. per-round communication cost (Qwen2-7B). Our methods are Pareto optimal, achieving the highest performance with the lowest communication cost.}
    \label{fig:perf_cost_time_qwen2_7b}
\end{figure}

\subsection{Experimental Results}

\subsubsection{Performance Comparison}
The main results are reported in Table~\ref{tab:main_results}. We analyze the results from three key perspectives.

\paragraph{In-Domain Performance.}
Across the experiments, our proposed methods, FedSRD and FedSRD-e, consistently achieve state-of-the-art performance on in-domain benchmarks. As shown in Table~\ref{tab:main_results}, with the Llama3.2-3B model, FedSRD and FedSRD-e achieve the highest average in-domain performance score, outperforming all baseline methods. This trend is further confirmed with the larger Qwen2-7B model, where FedSRD and FedSRD-e again lead with top performance, respectively. While some baselines may achieve peak performance on specific benchmarks, they often falter in other domains. In contrast, our framework exhibits robust and consistently high performance across all four specialized domains, underscoring its superior capability in aggregating diverse knowledge from specialist clients to build powerful global LLMs.

\paragraph{Communication Efficiency.}
A standout advantage of our framework is its remarkable communication efficiency. FedSRD and FedSRD-e reduce the per-round communication overhead to just 74 MB for Llama3.2-3B and 61 MB for Qwen2-7B, an approximate 90\% reduction compared to standard methods like FedAvg (742 MB) and advanced methods like FlexLoRA (742 MB). This is achieved by our proposed \textit{importance-aware sparsification} for uploads and \textit{alternative decomposition} for downloads (see Appendix~\ref{sec:comm_analysis} for a detailed analysis). Crucially, this significant gain in efficiency does not compromise performance. On the contrary, FedSRD and FedSRD-e achieve higher performance than baselines, directly answering our central research question. This is visualized in Figures~\ref{fig:perf_cost_time_llama3.2_3b} and~\ref{fig:perf_cost_time_qwen2_7b}, where our methods are shown to be Pareto optimal: achieving the best performance with the least communication.

\paragraph{Out-of-Domain Performance.}
Our methods also show strong performance on out-of-domain (OOD) benchmarks, effectively mitigating catastrophic forgetting. FedSRD and FedSRD-e achieve leading results on Llama3.2-3B and remain competitive on Qwen2-7B. Although FFA-LoRA exhibits strong OOD retention, we attribute this to its mechanism of freezing the LoRA $A$ matrices. which inherently constrains the capacity and hinders new knowledge acquisition in favor of preserving prior information. In contrast, our methods provide a more balanced and consistently high performance across both in-domain and out-of-domain evaluations. Further, we posit that the SVD projection in FedSRD, by seeking a low-rank consensus for in-domain tasks, may inadvertently filter out some general knowledge from the base model, which could account for the superior OOD performance of FedSRD-e. In summary, our framework not only excels at integrating specialized knowledge but also effectively preserves the general capabilities of the base model.

\subsubsection{Abation Study}
To validate each component of our FedSRD framework, we conducted an ablation study on Llama3.2-3B, with results in Table~\ref{tab:ablation_study}. Starting with the full model, we sequentially removed or replaced key components. Disabling the \textit{adaptive sparsity ratio} (using a fixed ratio instead) causes a slight performance drop, affirming the benefit of dynamic adjustment. Replacing our \textit{importance-aware sparsification} with a random baseline~\cite{yu2024language} leads to a significant degradation, highlighting that our decomposition method effectively preserves aggregated knowledge while reducing download communication costs. Further, ablating our \textit{alternative decomposition} (reverting to dense SVD download) also harms performance, confirming its effectiveness in preserving knowledge while reducing download costs. Finally, removing the reconstruction step (reverting to naive LoRA matrix averaging) results in severe performance loss, proving that full-rank aggregation is crucial for mitigating client heterogeneity. These results confirm that each component plays an indispensable and synergistic role in achieving the dual objectives of high model performance and extreme communication efficiency. Additionally, we show in Appendix~\ref{sec:plugin_fedavg} that our importance-aware sparsification also works as a more effective plugin to enhance conventional methods like FedAvg.

\begin{table}[tbp]
  \centering
  \caption{Ablation study on Llama3.2-3B. Components are removed sequentially. Note that \textit{importance-aware sparsification} is substituted with the random sparsification baseline~\cite{yu2024language} for its ablation.} 
  \label{tab:ablation_study} 
  \resizebox{\linewidth}{!}{
      \begin{tabular}{lcc} 
        \toprule
        \textbf{Method} & \textbf{In-Domain} & \textbf{OOD}  \\
        \midrule
        FedSRD                                 & 35.12 & 26.17 \\
        $-$ \textit{adaptive sparsity ratio} (Eq.~\ref{eq:kurtosis}) & 35.05 & 25.94 \\
        \quad  $-$ \textit{importance-aware sparsification} (Sec.~\ref{sec:importance_aware_sparse}) & 32.93 & 24.76 \\
        \quad \quad $-$ \textit{alternative decomposition} (Sec.~\ref{sec:alternative_decompose}) & 32.48 & 25.73 \\
        \quad \quad \quad $-$ \textit{reconstruct} (Sec.~\ref{sec:reconstruct}) & 32.32 & 23.90 \\ 
        \bottomrule
      \end{tabular}
  }
\end{table}

\begin{figure}[tbp]
    \centering
    \includegraphics[width=0.98\linewidth]{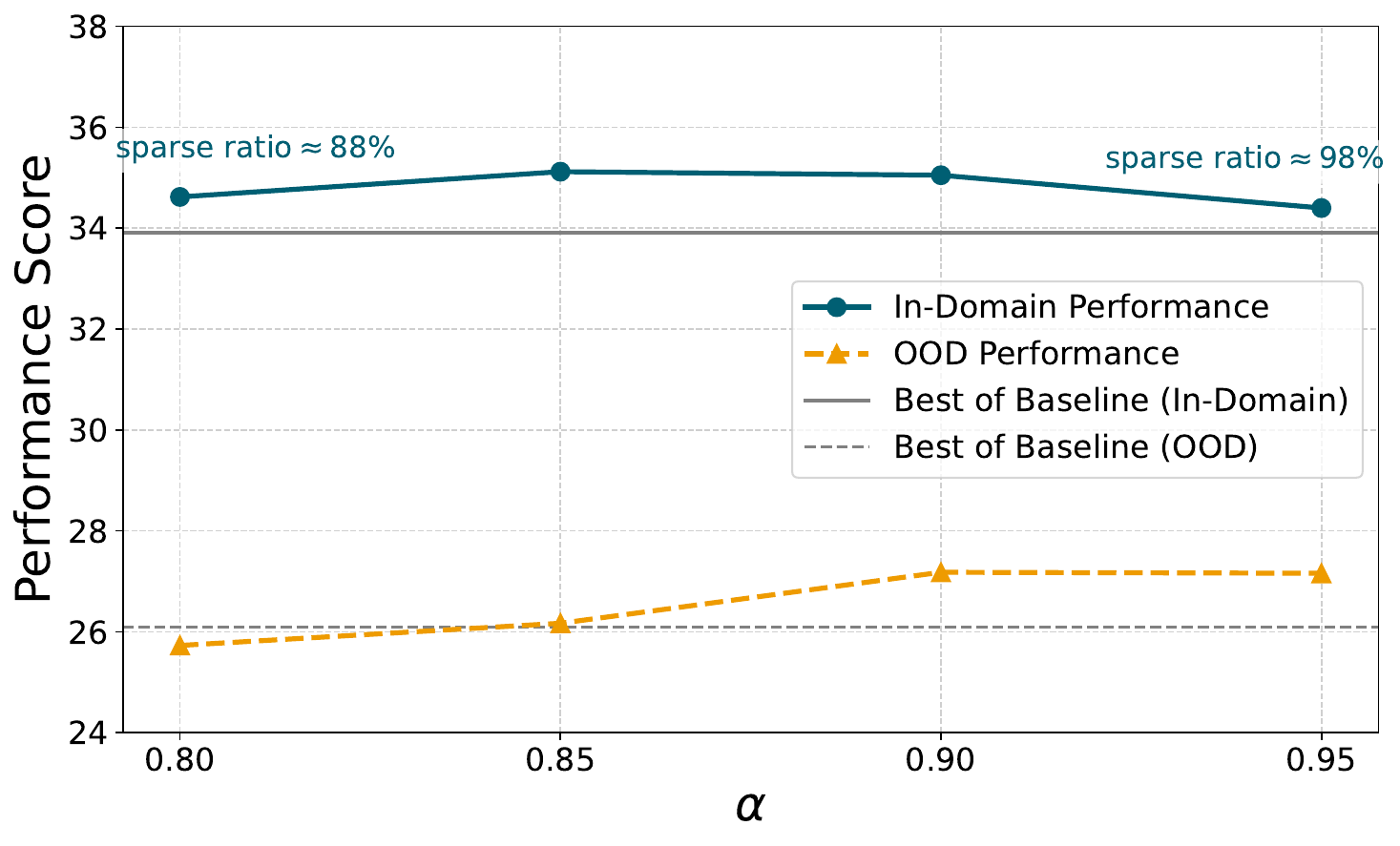}
    \caption{In-domain and out-of-domain (OOD) performance under different base sparsity values ($\alpha$) for FedSRD. Performance is robust across a range of sparsity levels, consistently outperforming the best baseline results.}
    \label{fig:alpha}
\end{figure}

\begin{table}[tbp]
\centering
\caption{In-domain and out-of-domain (OOD) performance under different download sparsification for FedSRD. Param. Reduction denotes the reduction ratio of the number of parameters downloaded per round.}
\label{tab:download_sparse}
\resizebox{\linewidth}{!}{
    \begin{tabular}{lccr}
    \toprule
    \textbf{Download Sparse Ratio} & \textbf{In-Domain} & \textbf{OOD} & \textbf{Param. Reduction} \\
    \midrule
    60\% & 34.32 & 26.76 & 80\% \\
    80\% & 35.12 & 26.17 & 90\% \\
    90\% & 34.57 & 25.13 & 95\% \\
    \bottomrule
    \end{tabular}
}
\end{table}

\subsubsection{Impact of Sparsification Ratios}

We first analyze the uplink base sparsity ratio, $\alpha$, in our importance-aware sparsification. As illustrated in Figure~\ref{fig:alpha}, FedSRD is robust and consistently outperforms all baselines even at an extreme sparsity of 98\% ($\alpha$=0.95). Interestingly, OOD performance improves with a higher $\alpha$. This suggests that uploading and aggregating fewer parameters helps preserve the pretrained model's generalizable knowledge by filtering out potentially conflicts in client-specific updates, thus preventing catastrophic forgetting. We can also observe that smaller $\alpha$ values (corresponding to lower sparsity) do not necessitate better performance, since preserving more parameters may introduce irrelevant or noisy knowledge for aggregation. A properly set $\alpha$ therefore offers a dual advantage: it significantly reduces communication costs while simultaneously reducing the potential interference for more effective aggregation.

For the downlink, we apply random sparsification to the decomposed matrices. Table~\ref{tab:download_sparse} shows that our method could even achieve better performance under aggressive sparsification. These results confirm that FedSRD achieves highly efficient communication and improved performance for federated LLMs fine-tuning.

\subsection{Computation Time Analysis}
We report the wall-clock time required for a single FL round in our simulation environment for both the Llama3.2-3B and Qwen2-7B in Table~\ref{tab:time_comparison_min}. Among the baselines, FedAvg provides a baseline for time consumption. FedProx requires slightly more time due to its proximal term, whereas FFA-LoRA is faster as it only trains and aggregates half of the LoRA matrices. FlexLoRA consumes more time performing SVD for matrix decomposition, and FedICU is the most time-intensive due to its computationally inefficient disentangled aggregation. For our methods, while the full FedSRD requires more time for its $r$-rank SVD projection, our efficient variant, \textbf{FedSRD-e, introduces negligible additional computation compared to FedAvg, while exhibiting performance second only to the full FedSRD}. This result highlights that our FedSRD-e variant achieves a Pareto-optimal balance across model performance, communication efficiency, and time consumption, making it a highly practical and scalable solution for real-world federated LLMs fine-tuning.

\begin{table}[tbp]
\centering
\caption{Computation time per round for different methods.}
\label{tab:time_comparison_min}
\resizebox{0.9\linewidth}{!}{%
\begin{tabular}{lrr}
\toprule
\textbf{Time per Round (min)} & \textbf{Llama3.2-3B} & \textbf{Qwen2-7B} \\ \hline
FedAvg & 1.92 & 3.06 \\
FedProx & 2.10 & 3.18 \\
FFA-LoRA & 1.86 & 3.00 \\
FlexLoRA & 7.20 & 12.60 \\
FedICU & 30.36 & 51.60 \\ 
Fed-DARE & 1.92 & 3.06 \\ \midrule
FedSRD & 7.92 & 13.50 \\
\textbf{FedSRD-e} & 2.10 & 3.15 \\ \bottomrule
\end{tabular}%
}
\end{table}

\section{Conclusion}
In this paper, we address the critical challenge of communication overhead in federated LoRA fine-tuning for LLMs. We propose FedSRD, a novel Sparsify-Reconstruct-Decompose framework for communication-efficient federated LLMs fine-tuning. Its core components include a structure-aware, importance-based sparsification strategy on the client side; a conflict-mitigating reconstruction and aggregation step in the full-rank space on the server side; and a novel alternating decomposition method to produce sparse global updates for efficient download. We also introduce FedSRD-e, a computationally light variant for resource-constrained environments. Through extensive experiments on diverse benchmarks, we demonstrated that FedSRD reduces communication costs by over 90\% while simultaneously improving model performance compared to state-of-the-art methods. Our work presents a practical and high-performing solution that significantly lowers the barrier for scalable, collaborative AI development on the decentralized Web.

\begin{acks}
This work was supported by the National Key R\&D Program of China under Grant No. 2022YFB2703301. This work was also supported by the Science and Technology Development Program of Two Districts in Xinjiang, China under Grant No. 2024LQ03004.
\end{acks}

\bibliographystyle{ACM-Reference-Format}
\balance
\bibliography{sample-base}

\appendix

\section{Pseudo-Code of FedSRD and FedSRD-e}
\label{sec:pseudo-code}

\begin{algorithm}[H]
\caption{FedSRD and FedSRD-e Framework}
\label{alg:fedsrd_full}
\begin{algorithmic}[1]
\STATE \textbf{Server Initializes:} Global LoRA matrices $B^0, A^0$. Let $\mathcal{C}$ be the set of all clients.

\FOR{each communication round $t = 1, 2, \dots, T$}
    \IF{$t > 1$}
        \STATE Server broadcasts $\Delta B_s^{t-1}$ or $\Delta A_s^{t-1}$.
        \FOR{each client $i \in \mathcal{C}$ \textbf{in parallel}}
            \STATE Client updates local model: $B_i^{t-1} \gets B_i^{t-2} + \Delta B_s^{t-1}$, $A_i^{t-1} \gets A_i^{t-2} + \Delta A_s^{t-1}$.
        \ENDFOR
    \ENDIF
    
    \FOR{each client $i \in \mathcal{C}$ \textbf{in parallel}} 
        \STATE $\Delta B_{i,s}^t, \Delta A_{i,s}^t \gets \text{ClientTraining}(i)$
    \ENDFOR
    
    \STATE \textbf{Server-Side Reconstruction and Aggregation:}
    \FOR{each client $i \in \mathcal{C}$}
        \STATE Client reconstructs its new state $B_i^t, A_i^t$ using Eq.~\ref{eq:lora_recover}
        \STATE Server computes client's full-rank matrix: $W_i^t \gets B_i^t A_i^t$.
    \ENDFOR
    \STATE Aggregate in full-rank space: $W^t \gets \frac{1}{|\mathcal{C}|} \sum_i W_i^t$.
    
    \STATE \textbf{Server-Side Alternating Decomposition:}
    \STATE $W^t_r \gets \text{SVD}_r(W^t)$. \COMMENT{FedSRD-e bypasses this SVD step.}
    \STATE Let $\Delta W^t_{update} = W^t_r - B^{t-1}A^{t-1}$. \COMMENT{For FedSRD-e, use $W^t$ instead of $W^t_r$.}
    \IF{$t$ is even}
        \STATE $\Delta A^t \gets (B^{t-1})^{\dagger} (\Delta W^t_{update})$
        \STATE $\Delta B^t \gets \mathbf{0}$
    \ELSE 
        \STATE $\Delta B^t \gets (\Delta W^t_{update})(A^{t-1})^{\dagger}$
        \STATE $\Delta A^t \gets \mathbf{0}$
    \ENDIF
    \STATE Update global state: $B^t \gets B^{t-1} + \Delta B^t$, $A^t \gets A^{t-1} + \Delta A^t$.
    \STATE Sparsify for download: $\Delta B_s^t, \Delta A_s^t \gets \text{DARE}(\Delta B^t, \Delta A^t)$.
\ENDFOR

\STATE \hrulefill
\STATE \textbf{Function} ClientTraining($i$):
\STATE \quad Fine-tune model on local data $\mathcal{D}_i$ to get new $B'_i, A'_i$.
\STATE \quad Compute local updates:$\Delta B_i^t \gets B'_i - B_i^{t-1}$, $\Delta A_i^t \gets A'_i - A_i^{t-1}$.
\STATE \quad \textbf{Importance-Aware Sparsification:}
\STATE \quad Calculate importance scores $I(\Delta B_i^t), I(\Delta A_i^t)$ using Eq.~\ref{eq:importance_score}.
\STATE \quad Calculate adaptive sparsity ratios $\rho_{\Delta B_i^t}, \rho_{\Delta A_i^t}$.
\STATE \quad Prune updates to get sparse $\Delta B_{i,s}^t, \Delta A_{i,s}^t$ adaptively.
\RETURN $\Delta B_{i,s}^t, \Delta A_{i,s}^t$ to server.
\end{algorithmic}
\end{algorithm}

\section{Rationale of the Importance Score Calculation}
\label{sec:rationale_imp}
We provide the mathematical motivation for the importance score defined in Equation~\ref{eq:importance_score}. The core idea is to quantify how much a single parameter's change in a LoRA update matrix ($\Delta B_i^t$ or $\Delta A_i^t$) contributes to the overall change in the full-rank weight matrix ($\Delta W_i^t$).

Let's first recall the expression of the weight update:
\begin{equation}
    \Delta W_i^t = \Delta B_i^t A_i^{t-1} + B_i^t \Delta A_i^t
    \label{eq:appendix_approx}
\end{equation}
We will analyze the contribution of a single element in $\Delta B_i^t$ to the term $\Delta B_i^t A_i^{t-1}$. The analysis for $\Delta A_i^t$ is symmetric.

A fundamental property of matrix multiplication is that the product of two matrices, $W=BA$, can be expressed as the sum of the outer products of the columns of $B$ and the rows of $A$. If $B \in \mathbb{R}^{d \times r}$ has columns $\mathbf{b}_j$ and $A \in \mathbb{R}^{r \times k}$ has rows $\mathbf{a}_j^T$, then:
\begin{equation}
    W = BA = \sum_{j=1}^{r} \mathbf{b}_j \otimes \mathbf{a}_j^T
\end{equation}
This expansion reveals the structural interplay between the two matrices: the final matrix $W$ is constructed from pairs of columns from $B$ and rows from $A$.

Now, let's consider the contribution of a single scalar element, $(\Delta B_i^t)_{uv}$ (the element in row $u$ and column $v$), to the matrix product $\Delta B_i^t A_i^{t-1}$. We can isolate this contribution by considering a change in only this single element. The effect of this single element on the product can be represented as:
\begin{equation}
    (\Delta B_i^t)_{uv} \cdot (\mathbf{e}_u \otimes \mathbf{e}_v^T) \cdot A_i^{t-1}
\end{equation}
where $\mathbf{e}_u$ and $\mathbf{e}_v$ are standard basis vectors (vectors of zeros with a single 1 at position $u$ or $v$, respectively). Using the associativity of matrix multiplication, this becomes:
\begin{equation}
    (\Delta B_i^t)_{uv} \cdot \mathbf{e}_u \otimes (\mathbf{e}_v^T A_i^{t-1})
\end{equation}
The term $\mathbf{e}_v^T A_i^{t-1}$ is a vector product that selects the $v$-th row of the matrix $A_i^{t-1}$, which we denote as $(A_i^{t-1})_{v,:}$. Thus, the contribution of the single element $(\Delta B_i^t)_{uv}$ to the final update matrix is itself a rank-1 matrix:
\begin{equation}
    \text{Contribution Matrix} = (\Delta B_i^t)_{uv} \cdot (\mathbf{e}_u \otimes (A_i^{t-1})_{v,:})
\end{equation}
This matrix is zero everywhere except for its $u$-th row, which is the $v$-th row of $A_i^{t-1}$ scaled by the scalar $(\Delta B_i^t)_{uv}$.

A natural way to measure the magnitude, or "importance", of this contribution matrix is to compute its Frobenius norm ($\|\cdot\|_F$). The Frobenius norm of a scaled outer product of two vectors $\mathbf{x}$ and $\mathbf{y}$ is $|c| \cdot \|\mathbf{x}\|_2 \cdot \|\mathbf{y}\|_2$. In our case, $c = (\Delta B_i^t)_{uv}$, $\mathbf{x} = \mathbf{e}_u$, and $\mathbf{y} = (A_i^{t-1})_{v,:}$. Since $\|\mathbf{e}_u\|_2 = 1$, the norm is:
\begin{equation}
    \begin{aligned}
    I((\Delta B_i^t)_{uv}) &= \left\| (\Delta B_i^t)_{uv} \cdot (\mathbf{e}_u \otimes (A_i^{t-1})_{v,:}) \right\|_F \\
    &= |(\Delta B_i^t)_{uv}| \cdot \|\mathbf{e}_u\|_2 \cdot \|(A_i^{t-1})_{v,:}\|_2 \\
    &= |(\Delta B_i^t)_{uv}| \cdot \|(A_i^{t-1})_{v,:}^T\|_2
\end{aligned}
\end{equation}

This is precisely the importance score for an element in $\Delta B_i^t$ as defined in Equation~\ref{eq:importance_score}. A symmetric derivation for the term $B_i^t \Delta A_i^t$ shows that the importance of $(\Delta A_i^t)_{uv}$ is its magnitude multiplied by the L2-norm of the $u$-th column of $B_i^t$. This formulation provides a principled way to measure a parameter's true structural contribution to the final update, moving beyond its simple magnitude.

\subsection{Theoretical Analysis of Pruning Error}
\label{sec:theoretical_analysis}

In this section, we provide a theoretical justification for the proposed importance score. We demonstrate that our greedy pruning strategy, based on the importance score defined in Equation~\ref{eq:importance_score}, is mathematically equivalent to minimizing the upper bound of the reconstruction error of the full-rank weight update $\Delta W$.

\subsubsection{Error Bound Derivation}
We analyze the approximation error for the term $\Delta W_B = \Delta B_i^t A_i^{t-1}$ (the derivation for $\Delta A$ is symmetric). Let $\mathcal{S}$ denote the set of indices in $\Delta B_i^t$ selected for pruning. The pruning error matrix $\mathcal{E}$ is the sum of the rank-1 contributions of the removed elements:
\begin{equation}
    \mathcal{E} = \sum_{(u,v) \in \mathcal{S}} (\Delta B_i^t)_{uv} \cdot (\mathbf{e}_u \otimes \mathbf{a}_v^T),
\end{equation}
where $\mathbf{a}_v^T$ is the $v$-th row of $A_i^{t-1}$. By applying the Triangle Inequality to the Frobenius norm of $\mathcal{E}$, we derive the error upper bound:
\begin{equation}
\begin{aligned}
    \|\mathcal{E}\|_F &= \left\| \sum_{(u,v) \in \mathcal{S}} (\Delta B_i^t)_{uv} (\mathbf{e}_u \otimes \mathbf{a}_v^T) \right\|_F \\
    &\leq \sum_{(u,v) \in \mathcal{S}} \underbrace{|(\Delta B_i^t)_{uv}| \cdot \|\mathbf{a}_v\|_2}_{I(\Delta B_i^t[u, v])}.
\end{aligned}
\label{eq:error_bound}
\end{equation}
The right-hand side is exactly the sum of the importance scores of the pruned parameters. Thus, pruning elements with the lowest scores $I(\cdot)$ strictly minimizes the upper bound of the weight update reconstruction error $\|\Delta W\|_F$.

\subsubsection{Comparison with Gradient-based Pruning}
Gradient-based pruning methods are impractical for Federated LoRA Fine-tuning for several reasons:

\begin{itemize}
    \item \textbf{Computational and Memory Overhead:} Calculating the exact influence of parameters on the loss usually requires second-order derivatives (e.g., Hessian). For LLMs, computing or even approximating the Hessian is prohibitively expensive for resource-constrained edge devices. Even first-order methods require retaining gradients, doubling the memory footprint during the scoring phase.
    \item \textbf{Federated Constraints:} In a federated setting, data is non-IID. A gradient-based score computed on local data may not accurately reflect the parameter's importance to the global model, potentially leading to overfitting local distributions.
\end{itemize}

In contrast, our proposed importance score is a \textit{data-free} and \textit{zero-overhead} metric. It relies solely on the magnitude of the weights and the structural correlations within the LoRA matrices ($\Delta B$ and $A$), which are already present in memory. This ensures that our method minimizes the deviation in the weight space ($\|\Delta W\|$) efficiently, offering a superior trade-off between communication efficiency and model performance.

\begin{table}[tbp]
\centering
\caption{Average In-domain Performance Comparison} 
\label{tab:sparse_comparison}
\resizebox{\linewidth}{!}{
\begin{tabular}{lr}
\toprule
 & Performance \\
\midrule
FedAvg & 33.90 \\
\quad with Random sparsification (i.e., Fed-DARE) & 33.57 \\
\quad with Magnitude sparsification & 31.61 \\
\quad with \textbf{Importance-aware sparsification (Our proposed)} & \textbf{34.04} \\
\bottomrule
\end{tabular}
}
\end{table}

\begin{table}[tbp]
\centering
\caption{Communication cost comparison per round (MB). Costs include both parameters and bitmaps where applicable.}
\label{tab:comm_cost}
\resizebox{\linewidth}{!}{
    \begin{tabular}{lrrr}
    \toprule
    \textbf{Method} & \textbf{Upload (MB)} & \textbf{Download (MB)} & \textbf{Total (MB)} \\
    \midrule
    FedAvg / FedProx / FlexLoRA & 371 & 371 & 742 \\
    FFA-LoRA & 185.5 & 185.5 & 371 \\
    FedICU & $\approx$197 & 371 & $\approx$568 \\
    Fed-DARE & $\approx$48.7 & $\approx$141.5 & $\approx$190 \\
    \textbf{Ours (FedSRD / FedSRD-e)} & $\approx$\textbf{31.0} & $\approx$\textbf{43.0} & $\approx$\textbf{74.0} \\
    \bottomrule
    \end{tabular}
}
\end{table}

\section{Communication Cost Analysis}
\label{sec:comm_analysis}

We analyze the communication cost per round for each method, using a Llama3.2-3B model fine-tuned with LoRA (rank $r=64$) as a case study. The model contains 97,255,424 LoRA parameters stored in float32 format. The baseline payload for transmitting the full, dense parameter set in one direction (upload or download) is $\frac{97,255,424 \times 4 \text{ bytes}}{1024^2 \text{ bytes/MB}} = 371$ MB. For sparse methods, an additional bitmap is required to encode the positions of non-zero parameters, costing $\frac{97,255,424 \text{ bits}}{8 \text{ bits/byte} \times 1024^2 \text{ bytes/MB}} \approx 11.6$ MB for the full set.

Table \ref{tab:comm_cost} details the communication costs for each method. Baseline approaches like FedAvg, FedProx, and FlexLoRA require a full round-trip transmission, totaling $742$ MB. FFA-LoRA halves this cost by communicating only the LoRA $B$ matrices. FedICU uses a selective upload ($\approx$50\% sparsity) and a dense download, for a total of $568$ MB. Fed-DARE, with a 90\% sparsity on uploads and an aggregated slightly more than 60\% sparsity on downloads, lowers the cost to $190$ MB. Our proposed method achieves the highest communication efficiency. Client-side importance-aware pruning results in a lean 31 MB upload. For download, the server transmits only half of the parameters (alternating between LoRA $A$ and $B$ matrices) with 80\% sparsity, leading to a download cost of just $43$ MB. This culminates in a total round-trip cost of only 74 MB, representing a 90\% reduction compared to FedAvg. The overhead from bitmaps can be further mitigated using lossless compression techniques like Run-Length Encoding~\cite{golomb1966run}.

\section{Importance-aware Sparsification Suits Well for FedAvg}
\label{sec:plugin_fedavg}
To validate the superiority of our proposed importance-aware sparsification over others, we plug and compare them on FedAvg in Table~\ref{tab:sparse_comparison}. The results show that our importance-aware sparsification is more effective in the FL system.

\section{Justification for Equation~\ref{eq:kurtosis}}
\label{sec:justification_eq7}
Equation~\ref{eq:kurtosis} proposes a data-driven approach to dynamically determine the layer-wise sparsity ratio, $\rho$, for the LoRA update matrices $\Delta A_i^t$ and $\Delta B_i^t$. The formulation is designed to adapt the pruning aggressiveness based on the intrinsic characteristics of each layer's importance score distribution, rather than relying on a uniform, manually-tuned sparsity level. The rationale behind each component of the equation is as follows:

\textbf{Kurtosis as a Heuristic for Prunability:} The core of our adaptive mechanism lies in using the kurtosis of the importance score distribution, $I(\cdot)$, as a quantitative indicator of a layer's "prunability." Kurtosis measures the "tailedness" of a distribution. A high kurtosis (a leptokurtic distribution) signifies that the importance scores are heavily concentrated on a small subset of parameters. This indicates a high degree of functional specialization and redundancy within the layer, making it an ideal candidate for aggressive pruning with minimal impact on performance. Conversely, a low kurtosis (a platykurtic distribution) implies that importance is more evenly diffused across many parameters. In this scenario, aggressive pruning would be detrimental, as it would likely remove parameters that make a non-trivial contribution. Therefore, a more conservative sparsity ratio is needed. By linking the sparsity ratio directly to kurtosis, our method automatically allocates a higher pruning budget to layers that can tolerate it and protects those that cannot.

\textbf{Logarithmic Scaling:} The raw values of kurtosis can span a wide range across different layers and training steps. We employ a logarithmic function to compress the wide range of kurtosis values into a more stable and manageable scale. This ensures that while a higher kurtosis leads to a higher sparsity ratio, the relationship is less sensitive to outlier kurtosis values, promoting more stable and robust pruning behavior throughout the process.

\textbf{Modulation Coefficient:} This coefficient scales the log-kurtosis term, modulating the final sparsity ratio. Empirically, we observe kurtosis values mostly ranging from 1 to 12 (with extremes up to 40). We selected a coefficient of 0.1. This choice is well-justified as it maps the typical kurtosis range to a moderate adjustment interval of approximately [0, 0.25]. This ensures the kurtosis contributes meaningfully to the sparsity calculation without introducing instability, making 0.1 a reasoned and effective choice.

\textbf{Base Sparsity Ratio ($\alpha$):} The hyperparameter $\alpha$ functions as a global base sparsity ratio. It establishes a baseline sparse level for the entire model, providing a direct and interpretable control over the overall sparsification ratio. The kurtosis-based term then acts as a dynamic, layer-specific offset to this baseline.

In summary, Equation~\ref{eq:kurtosis} implies the insight that the distribution of parameter importance reveals its tolerance to sparsifying. It begins with a global sparsity baseline ($\alpha$), which is then intelligently adjusted for each layer using a logarithmically scaled measure of its importance concentration (kurtosis), thereby achieving a robust, adaptive, and effective sparsity allocation strategy. We leave other possible designs for adaptive determination of sparsity ratio for future work.

\section{Magnitude of $B^{t-1}, A^{t-1}$ and $\Delta B^t, \Delta A^t$}
\label{sec:app_magnitude_ratio}
In the Taylor expansion in Equation~\ref{eq:taylor_expension}, the product term $\Delta B^t \Delta A^t$ is treated as a negligible higher-order term. To validate this assumption, we empirically analyze the magnitudes during the federated fine-tuning process. The average relative magnitude for matrix $A$, denoted as $\frac{\| \Delta A^t \|}{\| A^{t-1} \|}$, is less than 3\%, while for matrix $B$, $\frac{\| \Delta B^t \|}{\| B^{t-1} \|}$, it is less than 10\%. Their product constitutes a higher-order term that can be safely omitted. This conclusion is further reinforced by the observed convergence trend, where both relative ratios systematically decrease as the number of federated rounds increases.

\begin{table}[tbp]
\centering
\caption{Performance of base models and local training}
\label{tab:base_local_model_perf_reordered}
\begin{tabular}{l cc cc}
\toprule
& \multicolumn{2}{c}{Llama3.2-3B} & \multicolumn{2}{c}{Qwen2-7B} \\
\cmidrule(lr){2-3} \cmidrule(lr){4-5}
Model      & In-Domain & OOD     & In-Domain & OOD     \\
\midrule
Base       & 16.33     & 14.90   & 44.79     & 38.83   \\
Code       & 29.74     & 24.19   & 48.93     & 41.84   \\
Medical    & 25.49     & 23.65   & 52.11     & 40.92   \\
Finance    & 27.73     & 23.33   & 49.44     & 35.29   \\
Math       & 19.25     & 18.80   & 43.83     & 34.16   \\
\bottomrule
\end{tabular}
\end{table}

\section{Performance of Base Models and Locally Trained Models}
We test the performance of base models (Llama3.2-3B and Qwen2-7B) on our benchmarks. We also conduct local training on each client (e.g., Code indicates the model is locally trained on the client from the code domain) and test the locally trained models on each client on our benchmarks. The results are reported in Table~\ref{tab:base_local_model_perf_reordered}.

\end{document}